\documentclass[letterpaper]{article} 
\usepackage{aaai25} 
\usepackage{times}  
\usepackage{helvet}  
\usepackage{courier}  
\usepackage[hyphens]{url}  
\usepackage{graphicx} 
\usepackage{natbib}  
\usepackage{xcolor}
\usepackage{colortbl}
\usepackage{caption} 
\usepackage{makecell}  
\usepackage{booktabs}
\usepackage{amsmath} 
\usepackage{amssymb}
\usepackage{pifont}
\usepackage{multirow}

\usepackage[linesnumbered,ruled,vlined]{algorithm2e}
\SetKwInput{KwInput}{Input}     
\SetKwInput{KwOutput}{Output}     
\SetKwInput{KwFunction}{Function}

\renewcommand{\footnotesize}{\small}

\usepackage{tikz}
\usepackage{graphicx}
\usepackage{xcolor}

\definecolor{gray}{rgb}{0.78,0.7843,0.792}

\usepackage[capitalize]{cleveref}
\crefname{section}{Sec.}{Secs.}
\Crefname{section}{Section}{Sections}
\Crefname{table}{Table}{Tables}
\crefname{table}{Tab.}{Tabs.}
\DeclareCaptionStyle{ruled}{labelfont=normalfont,labelsep=colon,strut=off}  

\def\ie{\textit{i.e.}}
 
\def\eg{\textit{e.g.}}

\def\etal{et al.}

\usepackage{bibentry} 
\usepackage[absolute,overlay]{textpos}

\begin{document}

\title{Detecting and Corrupting Convolution-based Unlearnable Examples}

\author{ 
    Minghui Li\textsuperscript{\rm 1}\equalcontrib, 
    Xianlong Wang\textsuperscript{\rm 2,3}\equalcontrib\thanks{Corresponding author.}, 
    Zhifei Yu\textsuperscript{\rm 2}, Shengshan Hu\textsuperscript{\rm 2,3},\\
    Ziqi Zhou\textsuperscript{\rm 4}, 
    Longling Zhang\textsuperscript{\rm 2,3}, 
    Leo Yu Zhang\textsuperscript{\rm 5}
}
\affiliations{
    \textsuperscript{\rm 1} School of Software Engineering, Huazhong University of Science and Technology \\
    \textsuperscript{\rm 2} School of Cyber Science and Engineering, Huazhong University of Science and Technology \\
    \textsuperscript{\rm 3} Hubei Key Laboratory of Distributed System Security, Hubei Engineering Research Center on Big Data Security  \\
    \textsuperscript{\rm 4} School of Computer Science and Technology, Huazhong University of Science and Technology  \\
    \textsuperscript{\rm 5} School of Information and Communication Technology, Griffith University \\ 
    \{minghuili,wxl99,yzf,hushengshan,zhouziqi,longlingzhang\}@hust.edu.cn, 
    leo.zhang@griffith.edu.au \\    
}

\maketitle

\begin{abstract}
Convolution-based \textit{unlearnable examples} (UEs) employ class-wise multiplicative convolutional noise to training samples, severely compromising model performance. 
This fire-new type of UEs have successfully countered all defense mechanisms against UEs. 
The failure of such defenses can be attributed to the absence of norm constraints on convolutional noise, leading to severe blurring of image features. 
To address this, we first design an \underline{\textbf{E}}dge \underline{\textbf{P}}ixel-based \underline{\textbf{D}}etector (\textbf{EPD}) to identify convolution-based UEs. 
Upon detection of them, we propose the first defense scheme against convolution-based UEs, \underline{\textbf{CO}}rrupting these samples via random matrix multiplication by employing bilinear \underline{\textbf{IN}}terpolation (\textbf{COIN}) such that disrupting the distribution of class-wise multiplicative noise. 
To evaluate the generalization of our proposed COIN, we newly design two convolution-based UEs called VUDA and HUDA to expand the scope of  convolution-based UEs. 
Extensive experiments demonstrate the effectiveness of detection scheme EPD and that our defense COIN outperforms 11 \textit{state-of-the-art} (SOTA) defenses, achieving a significant improvement on the CIFAR and ImageNet datasets.
\end{abstract}

\section{Introduction} 

The triumph of \textit{deep neural networks} (DNNs) hinges on copious high-quality training data, motivating many commercial enterprises to scrape images from unidentified sources.  
In this scenario, adversaries may introduce elaborate and imperceptible perturbations to image data to serve as \textit{unlearnable examples} (UEs) that are subsequently disseminated online. 
This leads to a diminished generalization capacity of the victim model after being trained on such samples~\cite{unl,lin2024safeguarding,Meng2024SemanticDH}. 
Previous UEs apply additive perturbations under the $\mathcal{L}_p$ norm constraint to ensure sample's visual concealment, referred to as \textit{bounded UEs}~\cite{unl,fu2021robust,ren2023transferable,liu2024game}. 
Correspondingly, many defense schemes~\cite{bettersafe,liu2023image,pedro2023whatcanwe,wang2024eclipse} have been proposed, largely compromising bounded UEs. 
The ease with which bounded UEs are successfully defended can be attributed to the fact that the introduced additive noise is limited, rendering the noise distribution easily disrupted.

\newcommand{\cmark}{\ding{108}}  
\newcommand{\xmark}{\ding{109}}  
\begin{table}[t]
\centering
\scalebox{0.49}
{
\begin{tabular}{c|cccc}
\toprule[1.8pt]
\cellcolor[rgb]{.95,.95,.95}Defenses$\downarrow$  Metrics$\longrightarrow$ & 
\cellcolor[rgb]{.95,.95,.95}\makecell{Effectiveness \\ against \\ \textbf{CUDA} } & \cellcolor[rgb]{.95,.95,.95}\makecell{Effectiveness \\ against \\ \textbf{VUDA (Ours)} } & \cellcolor[rgb]{.95,.95,.95}\makecell{Effectiveness\\ against \\ \textbf{HUDA (Ours)} } & \cellcolor[rgb]{.95,.95,.95}\makecell{Model- \\ independent }  \\ \midrule[1.8pt]
Cutout~\cite{cutout} &\xmark   &\xmark   &\xmark   &\cmark   \\
Mixup~\cite{mixup} &\xmark   &\xmark   &\xmark   &\cmark   \\
Cutmix~\cite{cutmix} &\xmark   &\xmark   &\xmark   &\cmark   \\ 
DP-SGD~\cite{hong2020effectiveness} &\xmark   &\xmark   &\xmark   &\xmark   \\
AT~\cite{bettersafe}  &\xmark   &\xmark   &\xmark   &\xmark   \\
AA~\cite{qin2023learning}     &\xmark   &\xmark   &\xmark   &\cmark   \\
AVATAR~\cite{dolatabadi2023devil}     &\xmark   &\xmark   &\xmark   &\xmark   \\
OP~\cite{pedro2023whatcanwe}     &\xmark   &\xmark   &\xmark   &\xmark   \\
ISS-G~\cite{liu2023image}     &\xmark   &\xmark   &\xmark   &\cmark   \\
ISS-J~\cite{liu2023image}     &\xmark   &\xmark   &\xmark   &\cmark   \\
ECLIPSE~\cite{wang2024eclipse}  &\xmark   &\xmark   &\xmark   &\xmark   \\
\cellcolor[rgb]{.95,.95,.95}\textbf{COIN (Ours)}     &\cellcolor[rgb]{.95,.95,.95}\cmark  &\cellcolor[rgb]{.95,.95,.95}\cmark   &\cellcolor[rgb]{.95,.95,.95}\cmark   &\cellcolor[rgb]{.95,.95,.95}\cmark   \\ 
\bottomrule[1.8pt]
\end{tabular}
}
\caption{The effectiveness of existing defenses against three  convolution-based UEs and the dependence on models are presented. ``$\bullet$'' denotes fully satisfying the condition.}
\label{tab:related_work} 
\vspace{-5mm}
\end{table} 

\begin{figure*}[t]
\centering
\includegraphics[width=0.99\textwidth]{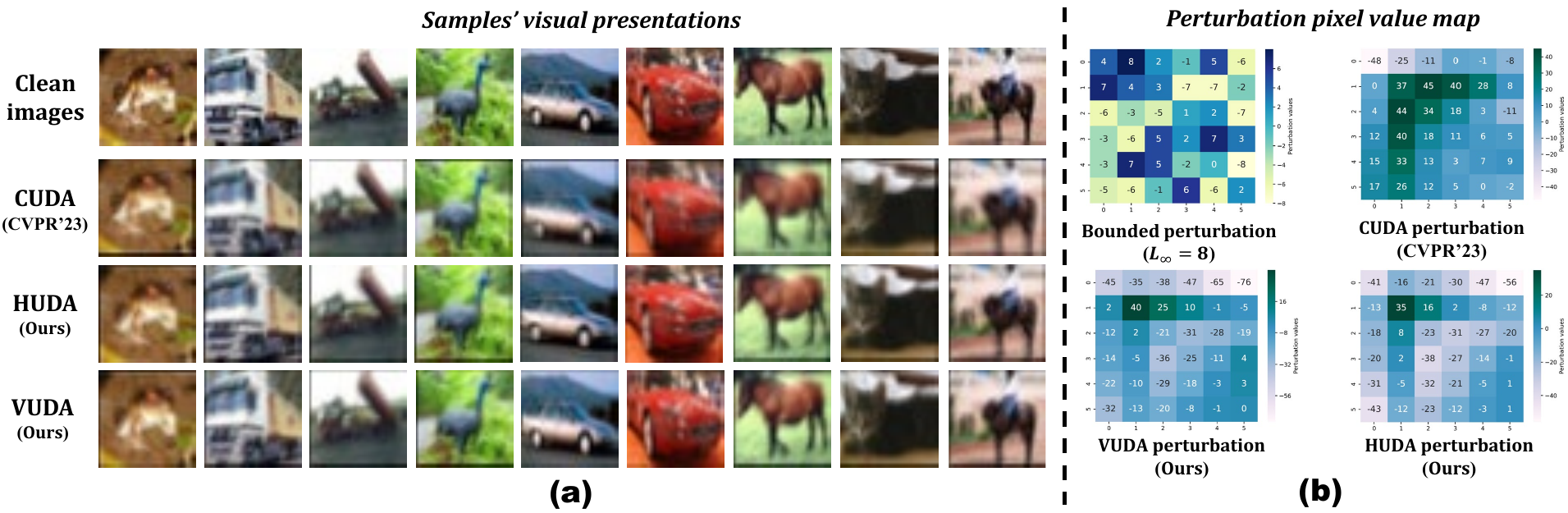}
\caption{(a) Visual images of clean samples and three  convolution-based UEs,  CUDA~\cite{sadasivan2023cuda}, HUDA, and VUDA. (b) The plots of perturbation values from bounded and convolution-based UEs. It can be seen that the bounded perturbation values are limited within a certain range, while convolution-based perturbations lack such constraints. }
\label{fig:vis-cuda-per} 
\vspace{-5mm}
\end{figure*}

However, the latest proposed UE that employs convolution operations without norm constraints, known as \textit{convolution-based UE}~\cite{sadasivan2023cuda}, expands the scope of noise as demonstrated in~\cref{fig:vis-cuda-per} (b), compromising the performance of current defenses, as demonstrated in~\cref{tab:related_work}. 
\begin{quote}
    \centering
    \emph{To the best of our knowledge, none of the existing defense mechanisms can effectively defend against convolution-based UEs.}
\end{quote}

To address this, the first step is to detect convolution-based UEs due to their visual concealment as shown in~\cref{fig:vis-cuda-per} (a).  
We observe that the edge pixel values of the convolution-based samples are biased towards black. 
Motivated by this key observation, we propose a detection scheme based on statistics of edge pixel values, successfully identifying convolution-based samples from UEs.

Subsequently, to simplify the challenging problem of defending against convolution-based UEs,  
we align with~\cite{javanmard2022precise,min2021curious} in regarding the image samples as column vectors generated by a \textit{Gaussian mixture model} (GMM)~\cite{reynolds2009gaussian}. 
In this manner, existing convolution-based UEs can be expressed as the product of a matrix and clean samples. This multiplicative matrix can be directly understood as \textit{multiplicative perturbations}, in contrast to the \textit{additive perturbations} from the bounded UEs. 
Considering that the reason behind the effectiveness of convolution-based UEs is that 
DNNs incorrectly establish the mapping between class-wise multiplicative noise\footnote{``Class-wise multiplicative noise" means that images from the same class use the same multiplicative convolution, while those from different classes use different ones.} and the ground truth labels~\cite{sadasivan2023cuda}. 
In light of this, our key intuition is to disrupt the distribution of class-wise multiplicative noise by increasing the inconsistency within classes and enhancing the consistency between classes.

Hence, we formally define two quantitative 
metrics in GMM,  the inconsistency within intra-class multiplicative matrix ($\Theta_{imi}$)  and the consistency within inter-class multiplicative matrix ($\Theta_{imc}$). 
Our defense goal is to design an operation that can simultaneously enhance both $\Theta_{imi}$ and $\Theta_{imc}$. 
Specifically, we leverage a uniform distribution to generate random values and shifts to construct a random matrix $\mathcal{A}_r$, subsequently applied to multiply the convolution-based UEs in the GMM scenario. 
After conducting empirical experiments, we verify that the random matrix we design effectively boosts both $\Theta_{imi}$ and $\Theta_{imc}$, demonstrating defensive capabilities against convolution-based UEs.

Based on these insights, 
we propose the first defense strategy for  countering convolution-based UEs, termed as \textbf{COIN},  which employs a randomly multiplicative image transformation as its mechanism. 
To verify the defense generalization of COIN against convolution-based UEs, we further propose two new types of convolution-based UEs, referred to as HUDA and VUDA as shown in~\cref{fig:vis-cuda-per}.   
Extensive experiments reveal that our proposed defense COIN significantly overwhelms all existing defense schemes, ranging from 19.17\%-44.63\% in accuracy on CIFAR-10 and CIFAR-100 datasets, while ACC exceeds 85\% and AUC surpasses  0.85 of proposed detection approach EPD. 
Our main contributions are summarized as:
\begin{itemize}
    \item \textbf{First Defense Against Convolution-based UEs.} To the best of our knowledge, we propose the first detection and defense approach for convolution-based UEs,  which utilizes an edge pixel value detection and a 
    random matrix multiplication transformation. 
    
    \item \textbf{Novel Convolution-based UEs.}
    We newly propose two convolution-based UEs, termed as VUDA and HUDA, to supply the scope of convolution-based UEs for better convincing defense evaluation. 
    
    \item \textbf{Experimental Evaluations.} Extensive experiments against convolution-based UEs including existing ones and those we newly propose on four benchmark datasets and six types of model architectures validate the effectiveness of our detection and defense strategy. 
\end{itemize}

\section{Preliminaries}
\subsection{Threat Model} 
The attacker crafts convolution-based UEs by employing class-wise convolutional kernel $\mathcal{K}_i$ 
to each image $x_i$ in the training set $\mathcal{D}_{c}$, thus causing the model $F$ with parameter $\theta$ trained on this dataset to generalize poorly to a clean test distribution $\mathcal{D}$~\cite{sadasivan2023cuda}. Therefore, the attacker formally expects to work out the following bi-level objective:

\begin{equation} 
\label{eq1}
\centering
\max  \underset{(x, y) \sim \mathcal{D}}{\mathbb{E}}\left[\mathcal{L}\left(F\left(x ; \theta_{p}\right), y\right)\right]
\end{equation}
\begin{equation}
\label{eq2}
 \text { \textit{s.t.} } \theta_{p} =  \underset{\theta}{\arg \min } \sum_{\left(x_{i}, y_{i}\right) \in  \mathcal{D}_{c}} \mathcal{L}\left(F\left( 
   x_i \otimes  \mathcal{K}_i     ; \theta\right), y_{i}\right)
\end{equation}
where $(x_{i}, y_{i})$ represents the clean data from $\mathcal{D}_{c}$, $\mathcal{L}$ is a loss function, 
\eg, cross-entropy loss, and $\otimes$ represents the convolution operation, while ensuring the modifications to $x_i$ are not excessive for preserving the concealment of the sample. 
As for defenders, in the absence of any knowledge of clean samples $x_i$, they aim to perform certain operations on UEs to achieve the opposite goal of~\cref{eq1}.
Bounded UEs~\cite{unl,syn,wu2022one,liu2024stable,wang2024eclipse} add norm-constrained additive noise to the sample $x_i$ in~\cref{eq2}.

\begin{figure*}[t]
\centering 
{\includegraphics[width=0.325\textwidth]{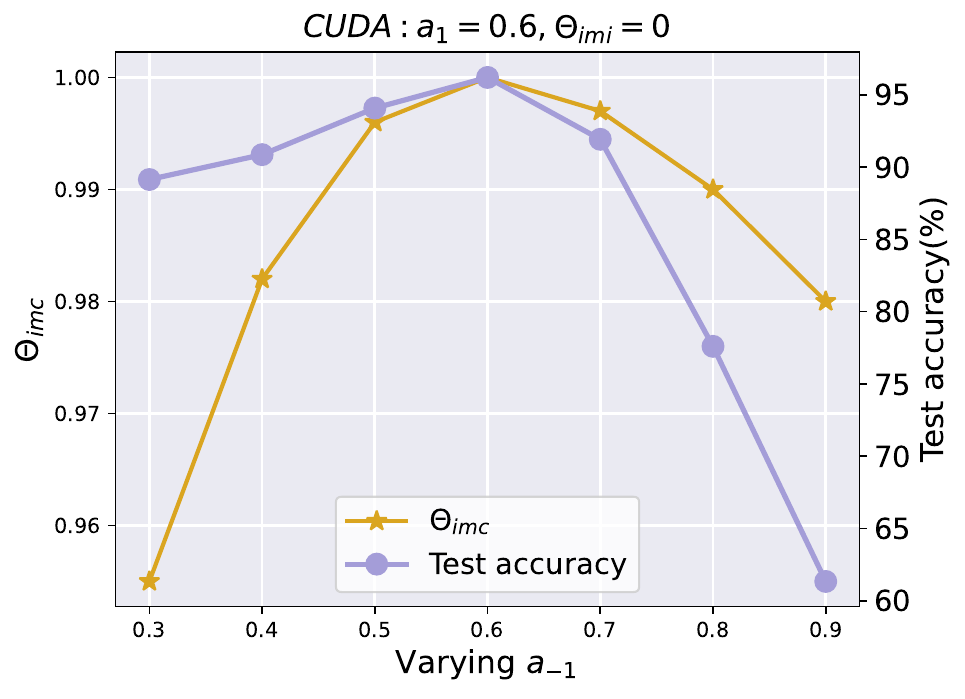}} 
{\includegraphics[width=0.325\textwidth]{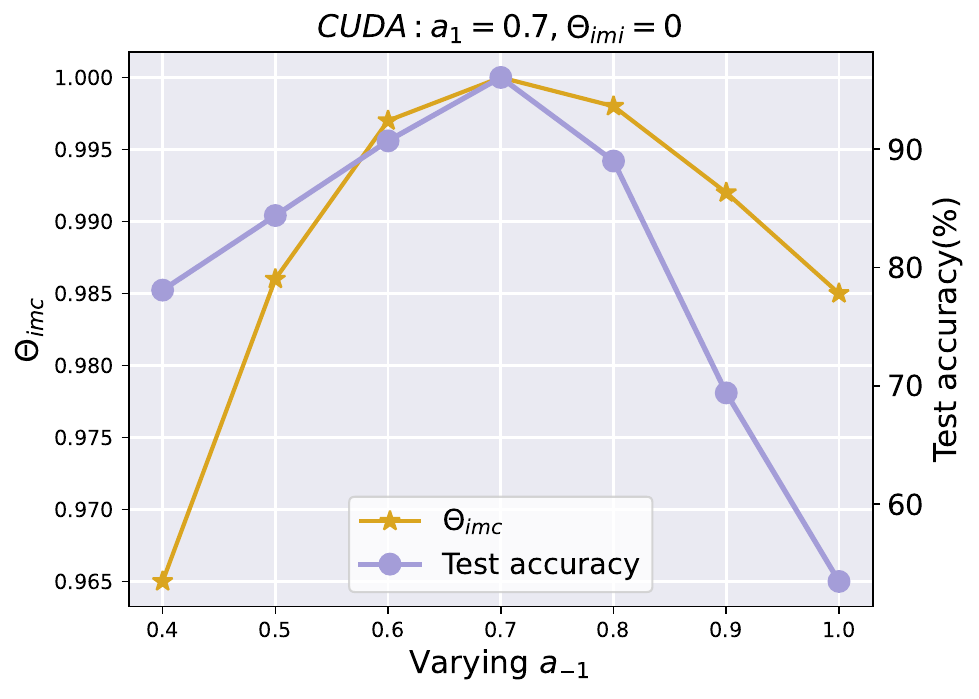}} 
{\includegraphics[width=0.325\textwidth]{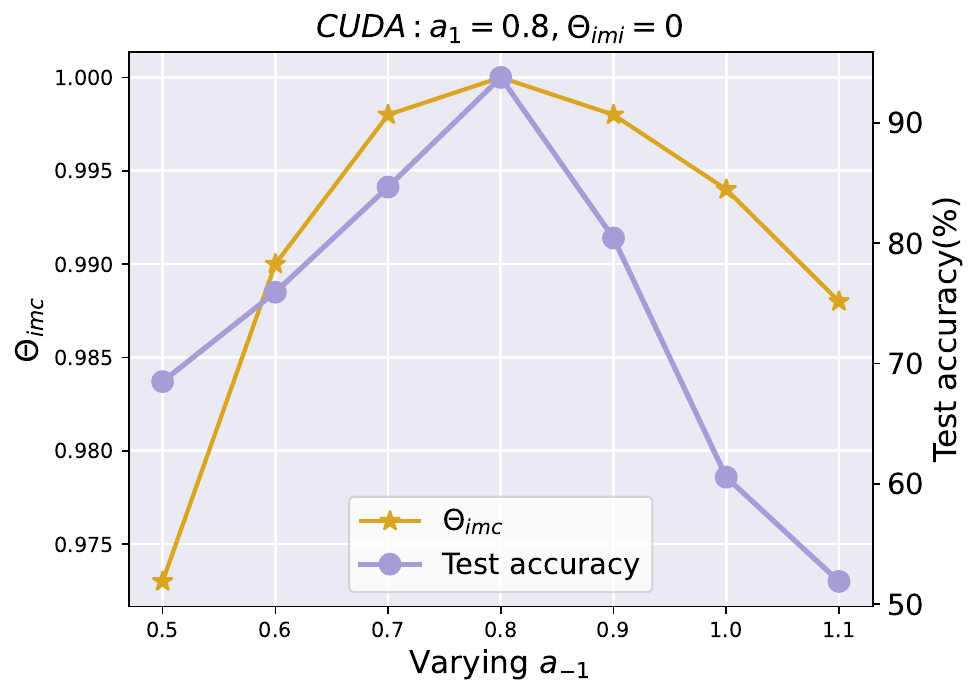}}
{\includegraphics[width=0.325\textwidth]{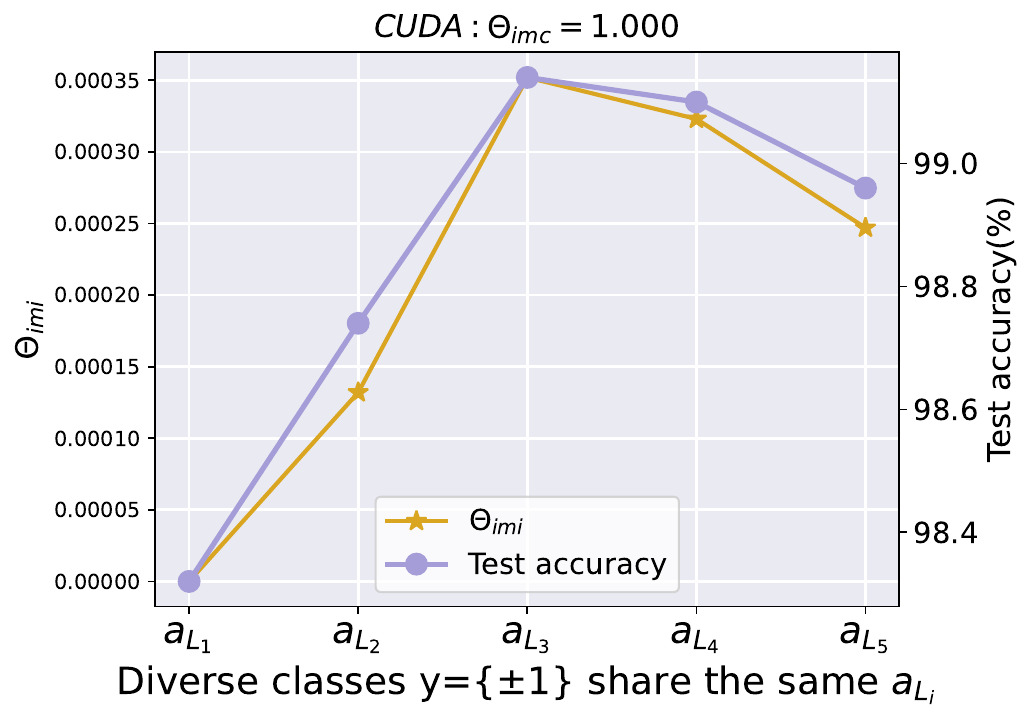}} 
{\includegraphics[width=0.325\textwidth]{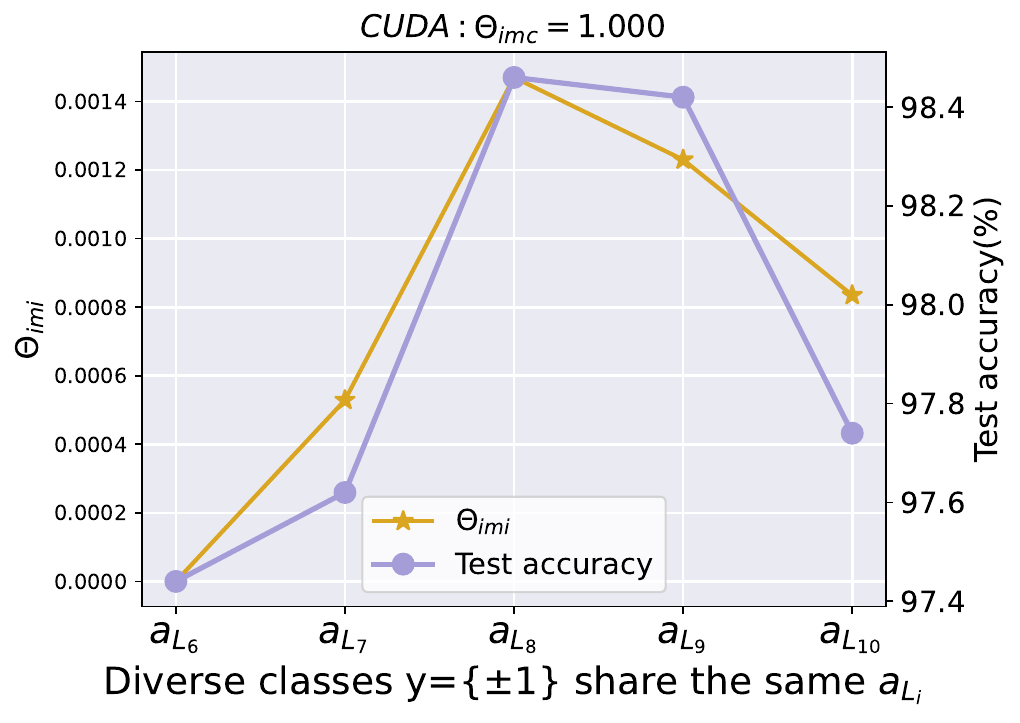}} 
{\includegraphics[width=0.33\textwidth]{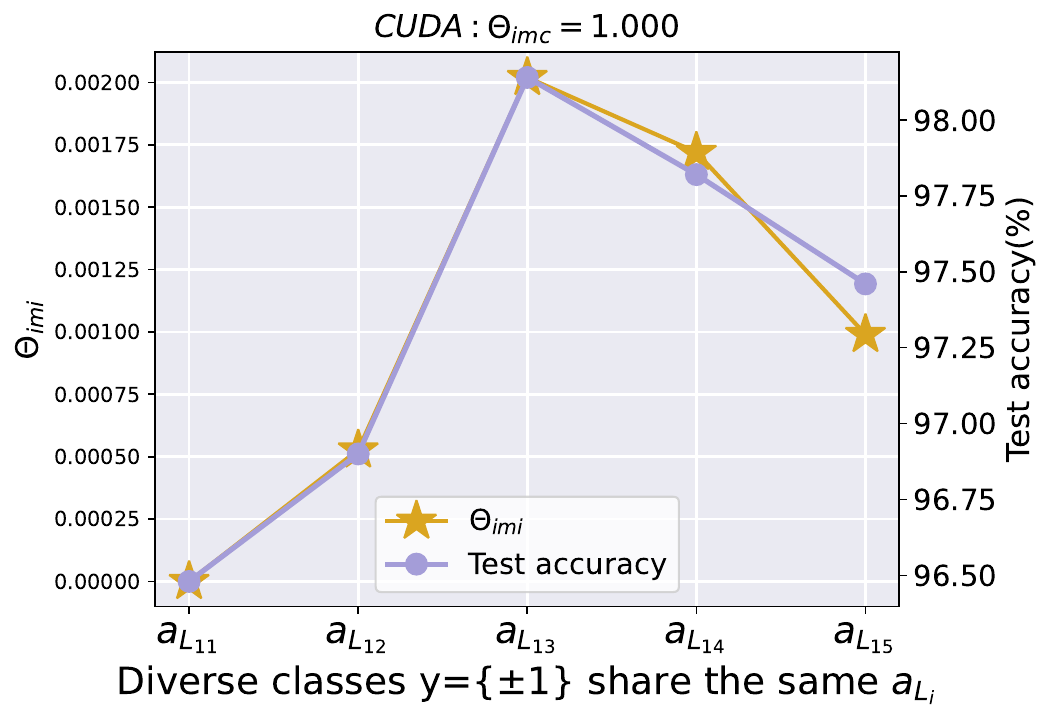}}
\caption{\textbf{Hypothesis validation.}  Test accuracy (\%)   with $\Theta_{imc}$ (top row) and $\Theta_{imi}$ (bottom row)  via changing parameter $a_y$.}
\label{fig3} 
\vspace{-3mm}
\end{figure*}

\subsection{Broadening the Attack Range}
Given the scarcity of research on existing convolution-based UEs, for more convincingly and comprehensively evaluating existing defense approaches against convolution-based UE attacks, we newly propose two types of convolution-based UEs, namely HUDA and VUDA, to expand the scope of convolution-based UEs. 
In particular, HUDA utilizes class-wise horizontal filters to apply convolutional operations, subsequently employed on images based on categories. The definition of HUDA's convolutional kernel is as follows:
\begin{equation}
    \mathcal{K}_h(i,j,k,l) = 
\begin{cases} b_y  & \text{ if } j= \frac{\mathcal{T}}{2} \quad and \quad i \in \{0,1,2\}\\
0  & \text{else}
\end{cases}
\end{equation}
where $i$ represents the channel, $j$ and $k$ denote the rows and columns of the convolutional kernel, and $l$ represents the depth of the convolutional kernel, $\mathcal{T}$ denotes the kernel size, and $b_y$ denotes the class-wise blur parameter. Similarly, the definition of the vertical convolutional kernel of VUDA is:
\begin{equation}
    \mathcal{K}_v(i,j,k,l) = 
\begin{cases} b_y  & \text{ if } k= \frac{\mathcal{T}}{2} \quad and \quad i \in \{0,1,2\}\\
0  & \text{else}
\end{cases}
\end{equation}
After obtaining the class-wise convolutional kernels, the entire dataset is processed using convolution operations to add the convolution-based multiplicative perturbations.

\subsection{Low-dimensional Representation}
\label{sec3.2}
\textbf{Decomposing in low-dimensional space.} Similar to~\cite{javanmard2022precise,min2021curious,wang2024unlearnable}, we define a binary classification problem involving a Bayesian classifier~\cite{friedman1997bayesian}, and the clean dataset $\mathcal{D}_{c}$ is sampled from a Gaussian mixture model $\mathcal{N}(y \boldsymbol{\mu}, I)$. Here, $y$ represents the labels $\{\pm 1\}$, with mean $\boldsymbol{\mu} \in \mathbb{R}^{d}$, and covariance $I \in \mathbb{R}^{d \times d}$ as the identity matrix ($d$ represents the feature dimension).
\begin{figure}[t]
\centering
\scalebox{0.47}{  
\includegraphics[width=\textwidth]{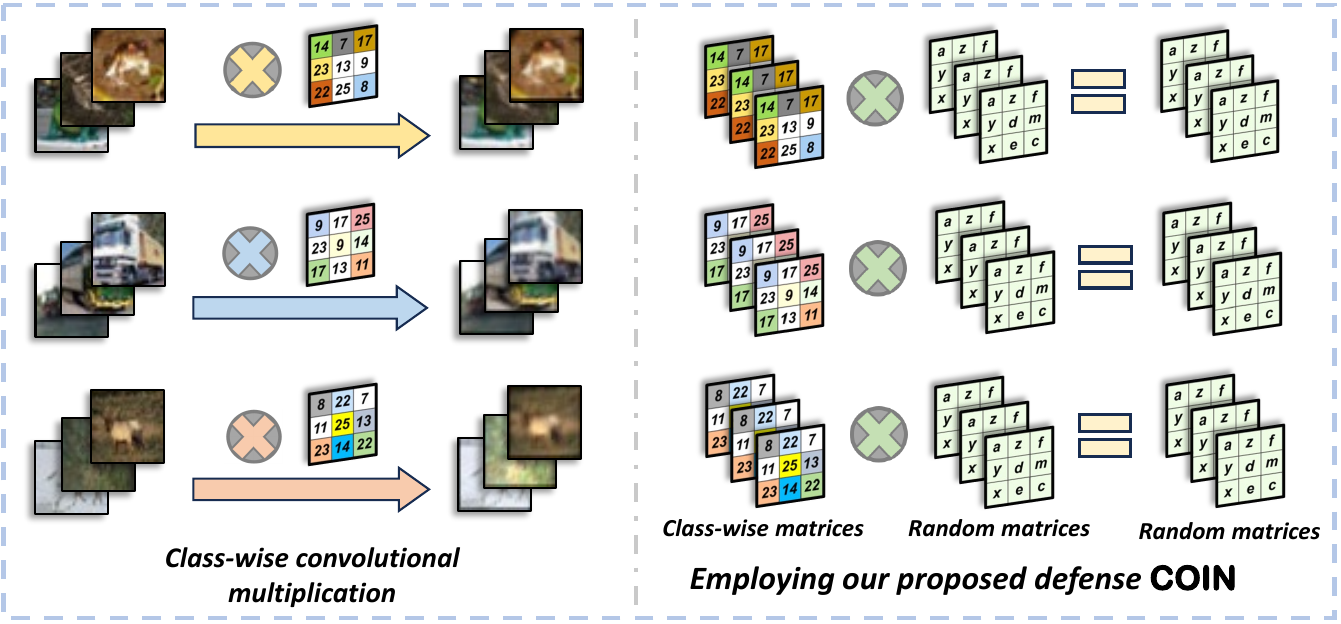}}
\caption{Key intuition behind our proposed defense.}
\label{fig:insight}  
\vspace{-3mm}
\end{figure}
We denote the clean sample as $x\in\mathbb{R}^{d}$, the convolution-based UE as $x_u$, which can be formulated as left-multiplying the class-wise matrices $\mathcal{A}_y$ by $x$:
\begin{equation}
\label{eq:key}
    x_u = \mathcal{A}_y\cdot x
\end{equation}
Specifically, CUDA employs a tridiagonal matrix $\mathcal{A}_{t}$, characterized by diagonal elements equal to 1, with the lower diagonal elements and upper diagonal elements both set to a pre-defined parameter $a_y$. 

\noindent\textbf{Key intuition.} 
The reason why convolution-based UEs work is that the model establishes a mapping between class-wise multiplicative noise and labels~\cite{sadasivan2023cuda}. Therefore, our intuition is that converting class-wise multiplicative noise into random multiplicative noise can disrupt this mapping, as demonstrated in~\cref{fig:insight}. 
Specifically, by increasing the inconsistency of intra-class multiplicative matrices or the consistency of inter-class multiplicative matrices, we can make the class-wise multiplicative noise information more disordered. 
This results in classifiers being unable to learn meaningless information from the multiplicative noise. 
\begin{figure*}[t]
\centering 
{\includegraphics[width=0.24 \textwidth]{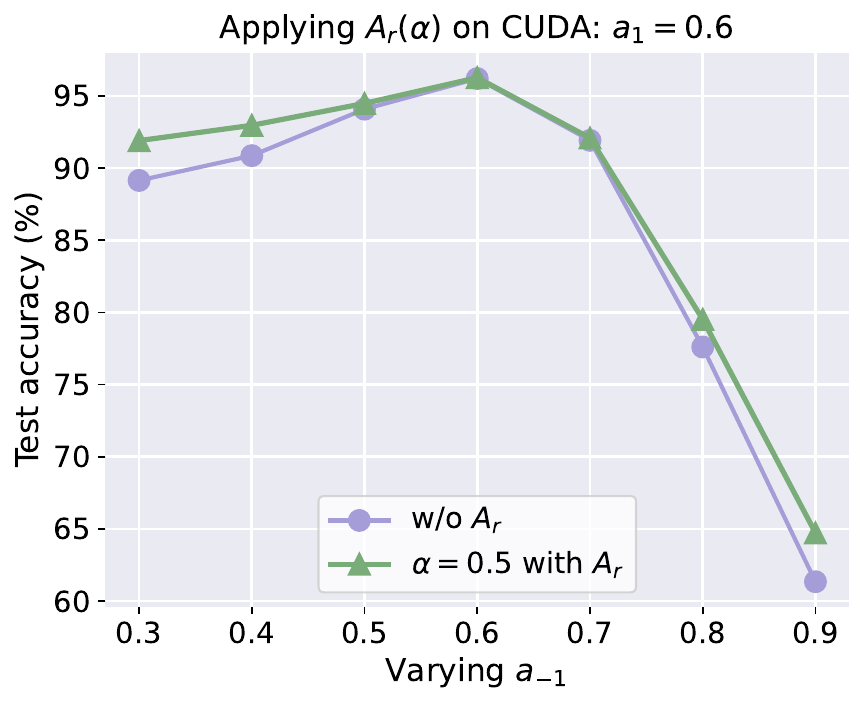}} 
{\includegraphics[width=0.24 \textwidth]{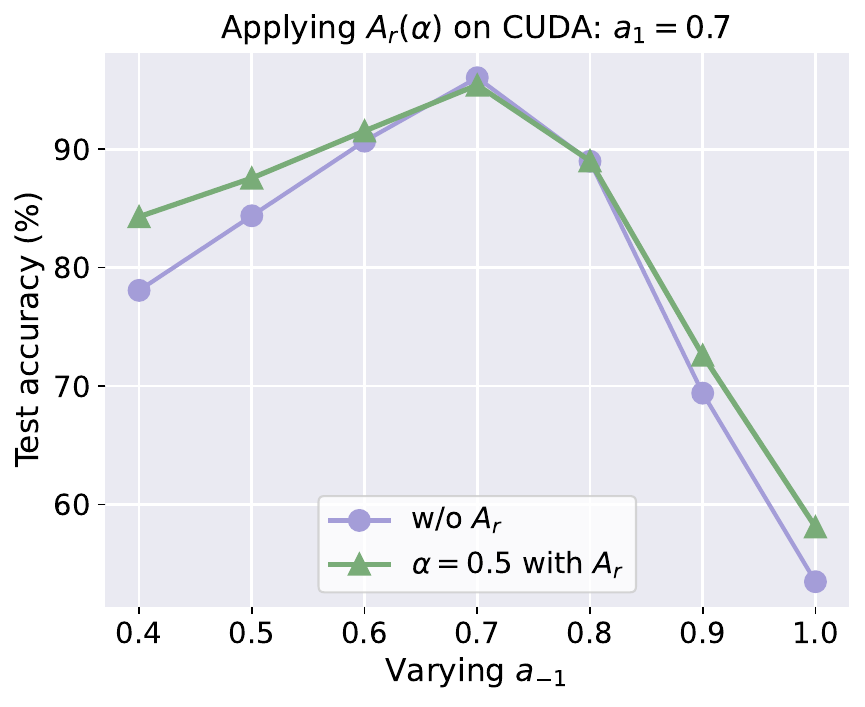}} 
{\includegraphics[width=0.24 \textwidth]{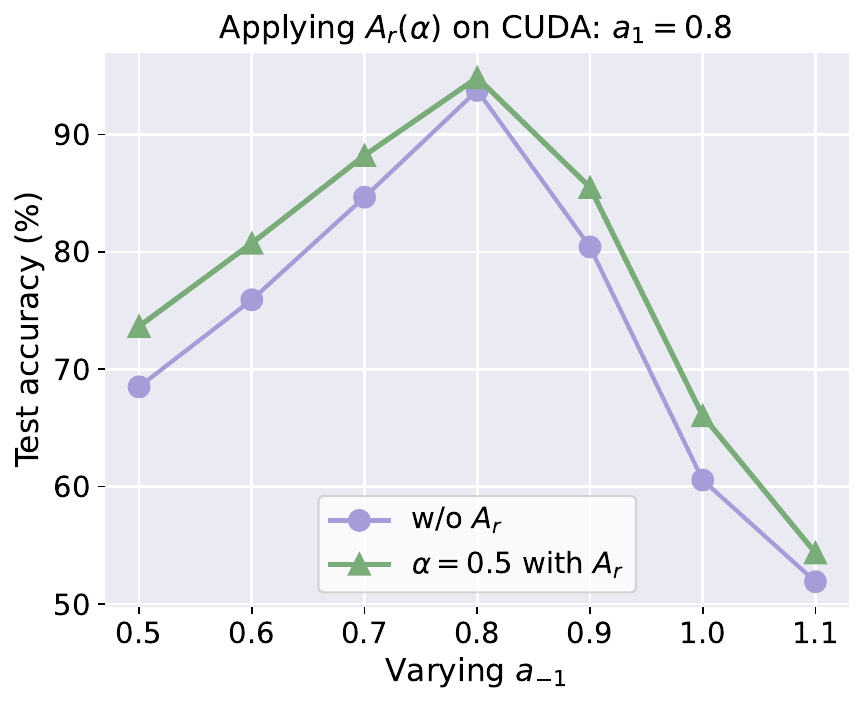}}
{\includegraphics[width=0.24 \textwidth]{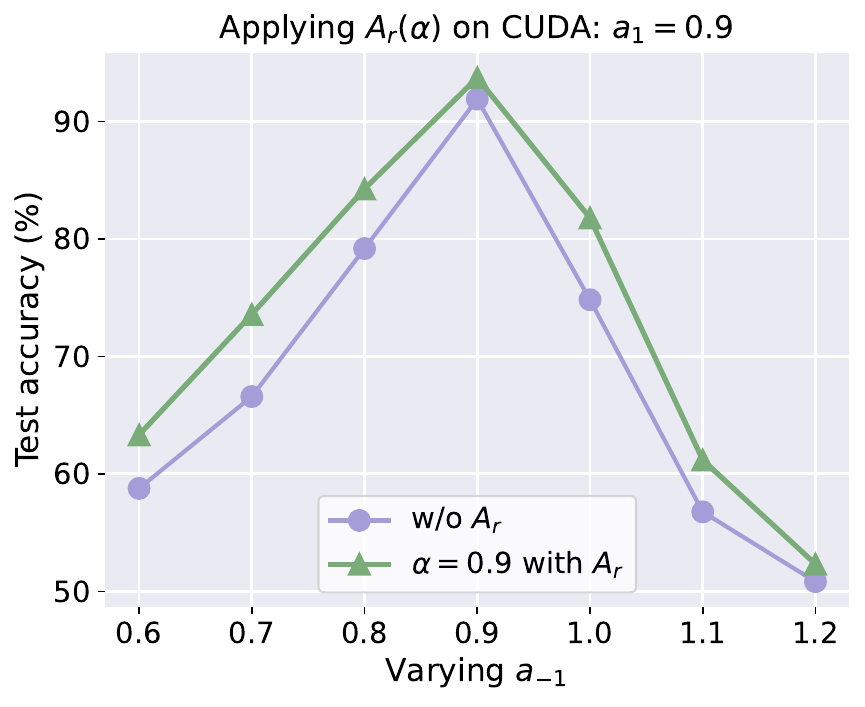}}
\caption{\textbf{Effectiveness of our proposed defense.} Comparison results of test accuracy before and after left-multiplying a random matrix $\mathcal{A}_r$ on all CUDA samples, $\alpha$ is set to 0.5.}
\label{fig:Ar_CUDA} 
\vspace{-3mm}
\end{figure*}
Based on this, we formally define the objective in the GMM space as follows:

\noindent\textbf{Definition 1.} \textit{We define the \textbf{\underline{i}ntra-class \underline{m}atrix \underline{i}nconsistency}, denoted as $\Theta_{imi}$, as follows: Given the multiplicative matrices $\{\mathcal{A}_i\mid i=1,2,\cdots,n\}$ within a certain class $y_k$ (containing $n$ samples), we have an intra-class average matrix defined as ${\mathcal{A}_\mu}_k =  \frac{1}{n} {\textstyle \sum_{i=1}^{n}} \mathcal{A}_i$, an intra-class matrix variance defined as $\mathbb{D}_k = \frac{1}{n} {\textstyle \sum_{i=1}^{n}}(\mathcal{A}_i - {\mathcal{A}_\mu}_k)^2 \in \mathbb{R}^{d \times d}$, an intra-class matrix variance mean value defined as ${\mathcal{V}_m}_k = \frac{1}{d^2}  \sum_{i=0}^{d-1}  \sum_{j=0}^{d-1} \mathbb D_k\left [ i \right ] \left [  j\right ] $, and then we have $\Theta_{imi}=\frac{1}{c} \sum_{k=0}^{c-1} {\mathcal{V}_m}_k$, where $c$ denotes the number of classes in $\mathcal{D}_{c}$.}

\noindent\textbf{Definition 2.} \textit{We define  the \textbf{\underline{i}nter-class \underline{m}atrix \underline{c}onsistency} as $\Theta_{imc}$, as follows: Given the flattened intra-class average matrices of the j-th and k-th classes $flat({\mathcal{A}_\mu}_j)$, $flat({\mathcal{A}_\mu}_k)$, 
we then have $\Theta_{imc} = sim (flat({\mathcal{A}_\mu}_j), flat({\mathcal{A}_\mu}_k))$, where $flat\left (  \right ) $ denotes flattening the matrix into a row vector and 
$sim(u,v)=uv^{T}/(\left \| u \right \|\left \| v\right \| ) $ denotes cosine similarity.}

\noindent\textbf{Hypothesis.} \textit{Increasing $\Theta_{imi}$ or $\Theta_{imc}$ can both improve the test accuracy of classifiers trained on the convolution-based UEs, and vice versa.}

\noindent\textbf{Empirical validation.} 
With the two metrics formalized and proposed hypothesis, we can conduct experiments to validate our key intuition.  
It can be observed from the top row of~\cref{fig3} ($\Theta_{imi}$ remains constant) that an increasing $\Theta_{imc}$ corresponds to an improvement in test accuracy, whereas a decrease in $\Theta_{imc}$ leads to a decline in test accuracy.  
In the bottom row of~\cref{fig3} ($\Theta_{imc}$ is held constant), test accuracy increases as $\Theta_{imi}$ rises and decreases as $\Theta_{imi}$ falls. 
Hence, these experimental results strongly support our proposed hypothesis and  key intuition.

\subsection{Design of Low-dimensional Defense}

Based on the results above, we are motivated to design a scheme that can increase $\Theta_{imi}$ and $\Theta_{imc}$, thus perturbing the distributions of multiplicative matrix $\mathcal{A}_y$. 
Assuming $\mathcal{A}_y$ is the most common diagonal matrix, we left-multiply 
$\mathcal{A}_y$ by a random matrix $\mathcal{A}_r \in \mathbb{R}^{d\times d}$ to disrupt it, where the diagonal of $\mathcal{A}_r$ is set to random values to introduce randomness.   
However, the form of diagonal matrix remains unchanged by multiplying this diagonal matrix, limiting the randomness.  
To solve this, we attempt to add another set of random variables above the diagonal, making it random. However, $\mathcal{A}_t$ employed in CUDA still maintains the tridiagonal matrix pattern, still contraining the randomness of the matrix. 
To introduce more randomness,  
we come up with the idea of introducing small random offsets to the random variables for each row, thus successfully breaking the tridiagonal form, without introducing additional variables.

Thus, we unify the random values and shifts via a uniform distribution $\mathcal{U}(-\alpha,\alpha)$, thereby striving to enhance $\Theta_{imc}$ as much as possible while already improving $\Theta_{imi}$. We first sample a variable $s \sim \mathcal{U}(-\alpha,\alpha), s \in \mathbb{R}^d$, and then obtain $m_i = \left \lfloor s_i \right \rfloor, n_i = s_i-m_i ,0 \le i \le d-1 $. $\mathcal{A}_r$ is designed as:
\begin{equation}
\label{eq:ar}
\resizebox{0.48\textwidth}{!}{$
\mathcal{A}_r  = 
\begin{bmatrix} 
  &\underline{1-n_0}   &\underline{n_0}  &0  &0 &\dots &0 \\
  &0  &\underline{1-n_1}  &\underline{n_1} &0 &\dots &0\\
  &0  &0  &\underline{1-n_2}   &\underline{n_2}  &\dots &0 \\
  &\vdots   &\vdots  &\vdots  &\ddots &\vdots &\vdots  \\
  &\vdots   &\vdots  &\vdots  &\vdots &\ddots &\vdots  \\
  &0  &\dots  &\dots  &\dots &0 &\underline{1-n_{d-1}} 
\end{bmatrix} \in \mathbb{R}^{d\times d}
$}
\end{equation}
where the $i$-th and ($i$+1)-th elements of the $i$-th row (0$\le i \le d$-1) are $1-n_i$ and $n_i$, and ``$\underline{1-n_i}, \underline{n_i}$" means that the positions of these two elements for each row are shifted by $m_i$ units simultaneously. When the new location $i+m_i$ or $i+1+m_i$ exceeds the matrix boundaries, we take its modulus with respect to $d$, thus obtaining their new position. The production process of random matrix $\mathcal{A}_r(\alpha)$ can be seen in the Alogorithm 1 in the supplementary material.

Whereupon, we obtain test accuracy results by left-multiplying all CUDA samples with the matrix $\mathcal{A}_r$, and compare with training on the CUDA UEs.
As shown in~\cref{fig:Ar_CUDA}, it can be observed that the test accuracy (with employing $\mathcal{A}_r$) is ahead of the accuracy (without using $\mathcal{A}_r$) regardless of values of $a_1$, demonstrating the defense effectiveness of our proposed random matrix $\mathcal{A}_r$. 

\section{Methodology} 
\subsection{Our Design for Defense Scheme: COIN}
\label{sec:method_coin}
Given the effectiveness of our defense scheme $\mathcal{A}_r$  in the low-dimensional space, our key intuition is that  
extending it to the real-image domain will successfully defend against convolution-based UEs. 
To achieve this, we note that the two random values in each row in~\cref{eq:ar} can be modeled as two weighting coefficients, while the random location offset $m_i$ in each row can be exploited to locate the pixel positions.  
Therefore, we regard the previous process of multiplying $\mathcal{A}_r$ as a random linear interpolation process, \ie, a random multiplicative transformation. 
In view of this, the convolution-based UE $x_u \in \mathbb{R}^{C\times H\times W}$ requires variables along with both horizontal and vertical directions and leveraging bilinear interpolation, which are defined as follows:
\begin{equation}
    s_x, s_y \sim \mathcal{U}(-\alpha, \alpha, size=H\cdot W)
\end{equation}
where $\mathcal{U}$ denotes a uniform distribution with size of $H\cdot W$ (height $\times$ width), $\alpha$ controls the range of the generated random variables. Considering that $s_x$ and $s_y$ are both arrays with size of $H\cdot W$, we obtain arrays $m_x$ and $m_y$ by rounding down each variable from the arrays to its integer part (\ie, random location offsets), formulated as: 
\begin{figure}[t]
\centering
\scalebox{0.47}{  
\includegraphics[width=\textwidth]{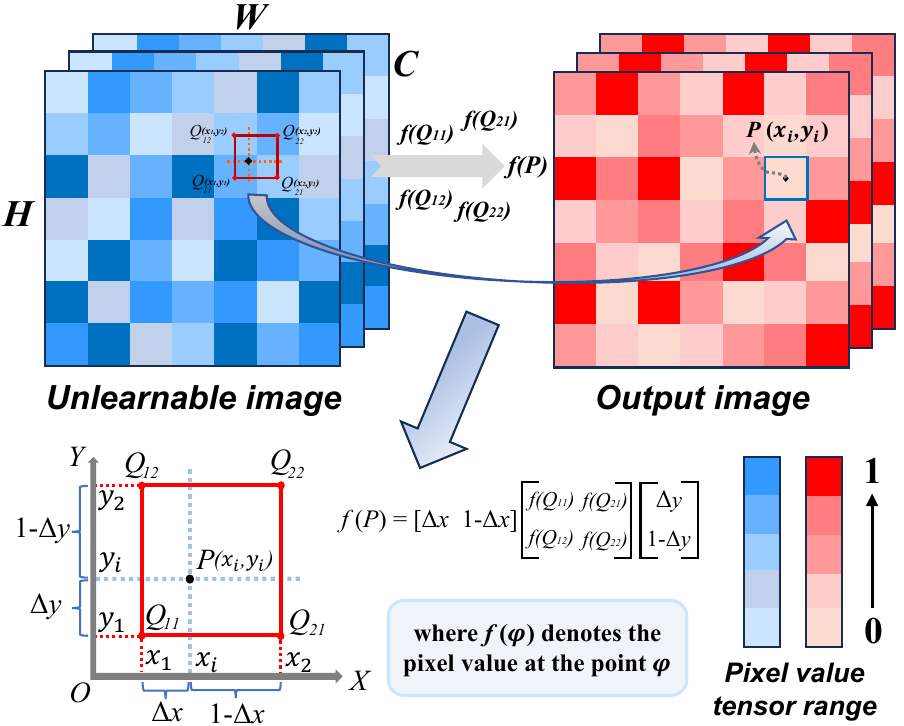}}
\caption{Our defense scheme COIN}
\label{fig:coin}  
\vspace{-3mm}
\end{figure}
\begin{equation}
    {m_x}_i =\left \lfloor  {s_x}_i \right \rfloor, \quad {m_y}_i =\left \lfloor  {s_y}_i \right \rfloor
\end{equation}
where $ \lfloor$  $ \rfloor $  represents the floor function, and $i$ is the index in the array, ranging from $0$ to $H\cdot W -1$. 
Subsequently, the arrays with coefficients $\omega_x$, $\omega_y$ required for the bilinear interpolation process are computed as follows:
\begin{equation}
    {\omega_x}_i = {s_x}_i-{m_x}_i, \quad {\omega_y}_i = {s_x}_i-{m_y}_i  
\end{equation}
To obtain the coordinates of the pixels used for interpolation, we first initialize a coordinate grid, which is defined as:
\begin{equation}
    c_x, c_y = \mathcal{M}(ara(W), ara(H)) \in \mathbb{R}^{H\times W}
\end{equation}
where $\mathcal{M}$ denotes coordinate grid creation function, $ara$ is employed to produce an array with values evenly distributed within a specified range. 
Whereupon we obtain the coordinates of the four nearest pixel points around the desired interpolation point in the bilinear interpolation process:
\begin{equation}
    {q_{11}}_i = (({c_x}_i + {m_x}_i)\%W, ({c_y}_i + {m_y}_i)\%H) 
\end{equation}
\begin{equation}
    {q_{21}}_i = (({c_x}_i + {m_x}_i+1)\%W, ({c_y}_i + {m_y}_i)\%H) 
\end{equation}
\begin{equation}
    {q_{12}}_i = (({c_x}_i + {m_x}_i)\%W, ({c_y}_i + {m_y}_i+1)\%H) 
\end{equation}
\begin{equation}
    {q_{22}}_i = (({c_x}_i + {m_x}_i+1)\%W, ({c_y}_i + {m_y}_i+1)\%H) 
\end{equation}
where $\%$ represents the modulo function, ensuring that the horizontal coordinate ranges from 0 to $W-1$ and the vertical coordinate ranges from 0 to $H-1$. 
Hence, we obtain new pixel values by using the pixel values of the four points mentioned above through bilinear interpolation:
\begin{equation}
\label{eq14}
 \mathcal{F}_j({p}_i)=\begin{bmatrix}
    {1 -\omega_x}_i & {\omega_x}_i\\
    \end{bmatrix}
    \begin{bmatrix}
     \mathcal{F}_j({q_{11}}_i) & \mathcal{F}_j({q_{12}}_i)\\
    \mathcal{F}_j({q_{21}}_i)  & \mathcal{F}_j({q_{22}}_i)
    \end{bmatrix}
    \begin{bmatrix}
     {1-\omega_y}_i\\
    {\omega_y}_i
    \end{bmatrix}
\end{equation}
where $\mathcal{F}_j()$ denotes the pixel value of the $j$-th channel at a certain coordinate point, $p_i$ denotes the coordinate of the newly generated pixel point. Each generated pixel value should be clipped within the range (0,1).
Finally, we gain the transformed image $x_t$ by applying~\cref{eq14} and clipping to pixel values of each channel of the image $x_u$. The schematic diagram of the above process is shown in~\cref{fig:coin}. 



\subsection{Our Design for Detection Scheme: EPD}
\textbf{Key Intuition.} It is crucial to detect whether the perturbations are convolution-based or not, because without prior knowledge, defenders have difficulty discerning the type of UEs.    
The key intuition behind our detection method is that convolution-based UEs typically exhibit lower pixel values around the edges, often appearing as black pixels as shown in~\cref{fig:vis-cuda-per} (a). 
Leveraging this intuition, we propose \textit{edge pixel-based detection} (EPD) scheme as follows:

\noindent\textbf{Feature Extraction.} We sum the RGB values along the four edges (\ie, top, bottom, left, and right) of the detection-ready sample $x_d \in \mathbb{R}^{C \times H \times W}$  to serve as the feature for detection. 
The channel-wise feature vector $v_d(c)$ is defined as:

\begin{equation}
\scriptsize
v_d(c) = \left[\sum_{w=1}^{W} e(c, 1, w), \sum_{w=1}^{W} e(c, H, w), \sum_{h=1}^{H} e(c, h, 1), \sum_{h=1}^{H} e(c, h, W)\right]
\end{equation}
where $e(c, h, w)$ represents the pixel value at channel $c$, row $h$, and column $w$. 
The final feature vector $\mathcal{V}_d$ is calculated by concatenating $v_d(c)$ across three channels, formulated as:
\begin{equation}
    \mathcal{V}_d = [v_d({R}), v_d({G}), v_d({B})]
\end{equation}
 

\noindent\textbf{Binary Classification for Detection.} 
Given the simplicity of the  vector $\mathcal{V}_d$, we utilize a \textit{support vector machine} (SVM)~\cite{huang2018applications} for binary classification. 
The pre-trained SVM classifies the image $x_d$  by comparing the decision score $\mathcal{S}$ with a pre-defined threshold $\theta$ (typically set to 0 in SVM), where  $\mathcal{S}$ is defined as: 
\begin{equation}
\label{eq:svm}
\mathcal{S} = \mathbf{w} \cdot \mathcal{V}_d + b
\end{equation}
where $\mathbf{w}$ is the weight vector of the SVM, and $b$ is the bias term. The predicted label $\hat{y} \in \{0, 1\}$ is given by:
\begin{equation}
\label{eq:predicty}
\hat{y} = 
\begin{cases} 
1 & \text{if } S(x_d) \geq \theta \\ 
0 & \text{if } S(x_d) < \theta 
\end{cases}
\end{equation}
where 1 denotes a convolution-based UE, while 0 does not. 
It is important to note that since our proposed defense COIN is designed specifically for convolutional UEs, we only apply COIN  after our EPD identifies the sample as a convolutional UE. 
Otherwise, whether the sample is a bounded UE, a clean example, or any other type, we do not consider differentiation or processing.

\begin{table*}[t]
\centering
\scalebox{0.67}{\begin{tabular}{c|ccccc|ccccc}
\toprule[1.8pt]
\cellcolor[rgb]{.95,.95,.95} Datasets 
&  \multicolumn{5}{c|}{\cellcolor[rgb]{.95,.95,.95} CIFAR-10} 
& \multicolumn{5}{c}{\cellcolor[rgb]{.95,.95,.95} CIFAR-100} \\
 \multicolumn{1}{l|}{\cellcolor[rgb]{.95,.95,.95} $\quad\quad\quad$Defenses$\downarrow\quad\quad\quad$Models$\rightarrow$} & \cellcolor[rgb]{.95,.95,.95}ResNet18       & \cellcolor[rgb]{.95,.95,.95}VGG16          & \cellcolor[rgb]{.95,.95,.95}DenseNet121    & \cellcolor[rgb]{.95,.95,.95}MobileNetV2    & \cellcolor[rgb]{.95,.95,.95}\textbf{AVG}            & \cellcolor[rgb]{.95,.95,.95}ResNet18       & \cellcolor[rgb]{.95,.95,.95}VGG16          & \cellcolor[rgb]{.95,.95,.95}DenseNet121    & \cellcolor[rgb]{.95,.95,.95}MobileNetV2    & \cellcolor[rgb]{.95,.95,.95}\textbf{AVG}  \\ \midrule[1.8pt]
w/o                                                        & 26.49          & 24.65          & 27.21          & 21.34          & 24.92          & 14.31          & 12.53          & 13.90          & 12.94          & 13.42          \\
MU~\cite{mixup}                                                         & 26.72          & 28.07          & 24.67          & 24.63          & 26.02          & 17.09          & 13.35          & 19.97          & 13.55          & 15.99          \\
CM~\cite{cutmix}                                                         & 26.02          & 28.53          & 24.64          & 20.73          & 24.98          & 12.51          & 10.14          & 20.77          & 10.14          & 13.39          \\
CO~\cite{cutout}                                                         & 20.07          & 27.58          & 24.86          & 20.46          & 23.24          & 12.80          & 10.56          & 16.19          & 13.56          & 13.28          \\
DP-SGD~\cite{hong2020effectiveness}                                                     & 25.50          & 23.02          & 25.25          & 25.78          & 24.89          & 12.42          & 10.56          & 16.36          & 12.72          & 13.02          \\
AVATAR~\cite{dolatabadi2023devil}                                                     & 30.67          & 29.57          & 33.15          & 28.53          & 30.48          & 14.49          & 10.81          & 12.97          & 13.85          & 13.03          \\
AA~\cite{qin2023learning}                                                         & 39.85          & 38.68          & 38.92          & 41.06          & 39.63          & 24.83          & 1.00           & 27.89          & 20.49          & 18.55          \\
OP~\cite{pedro2023whatcanwe}                                                         & 29.77          & 30.33          & 33.82          & 28.86          & 30.70         & 20.17          & 14.59          & 15.55          & 23.02          & 18.33          \\
ISS-G~\cite{liu2023image}                                                      & 25.77          & 21.42          & 26.73          & 19.85          & 23.44          & 8.80           & 6.40           & 11.48          & 8.71           & 8.85           \\
ISS-J~\cite{liu2023image}                                                      & 45.10          & 40.26          & 39.79          & 41.46          & 41.65          & 33.62          & 26.92          & 28.94          & 31.23          & 30.18          \\
AT~\cite{bettersafe}                                                         & 50.59          & 45.95          & 49.01          & 42.59          & 47.04          & 37.27          & 28.18          & 34.21          & 35.74          & 33.85          \\
ECLIPSE~\cite{wang2024eclipse}                                                         & 32.87          & 30.82          & 32.26          & 32.69          & 32.16          & 18.54          & 15.59          & 19.24          & 20.23        & 18.40          \\
\textbf{COIN (Ours)}                                       &\cellcolor[rgb]{0.686, 0.67, 0.67} \textbf{71.90} &\cellcolor[rgb]{0.686, 0.67, 0.67} \textbf{73.65} &\cellcolor[rgb]{0.686, 0.67, 0.67} \textbf{70.45} &\cellcolor[rgb]{0.686, 0.67, 0.67} \textbf{73.63} &\cellcolor[rgb]{0.686, 0.67, 0.67} \textbf{72.41} &\cellcolor[rgb]{0.686, 0.67, 0.67} \textbf{48.63} &\cellcolor[rgb]{0.686, 0.67, 0.67} \textbf{46.74} &\cellcolor[rgb]{0.686, 0.67, 0.67} \textbf{45.72} &\cellcolor[rgb]{0.686, 0.67, 0.67} \textbf{48.53} &\cellcolor[rgb]{0.686, 0.67, 0.67} \textbf{47.41} \\ \bottomrule[1.8pt]
\end{tabular}}
\caption{\textbf{Defenses against CUDA.} The \textit{test accuracy}  (\%) results  on CIFAR-10 and CIFAR-100 datasets.}
\label{tab:main} 
\vspace{-4mm}
\end{table*}

\section{Experiments} 
\subsection{Experimental Settings}
Networks including ResNet (RN)~\cite{resnet}, DenseNet (DN)~\cite{densenet}, MobileNetV2 (MNV2)~\cite{sandler2018mobilenetv2},  and VGG~\cite{vgg}  are selected. 
Benchmark datasets across from  small, medium, and large sizes like 
CIFAR-10 (32×32), CIFAR-100 (32×32)~\cite{cifar10}, ImageNet20 (64×64), and ImageNet100 (224×224)~\cite{imagenet} are used. 
The uniform distribution range $\alpha$ of COIN is empirically set to 2.0. During pre-training the SVM classifier of EPD, the linear kernel function is utilized, and penalty parameter $C_p$ is set to 0.1. Meanwhile, the threshold $\theta$ of EPD during detection is set to 0.  
During training of classification task, we use SGD for training 80 epochs with a momentum of 0.9, a learning rate of 0.1, and batch sizes of 128, 32 for CIFAR and ImageNet, respectively\footnote{Our code is available at \url{https://github.com/wxldragon/COIN}.}.

\begin{table}[t] 
\centering
\scalebox{0.49}{
\begin{tabular}{c|ccccc|ccccc}
\toprule[1.8pt]
\cellcolor[rgb]{.95,.95,.95} Datasets  & \multicolumn{5}{c|}{\cellcolor[rgb]{.95,.95,.95}ImageNet100 224×224} & \multicolumn{5}{c}{\cellcolor[rgb]{.95,.95,.95}ImageNet20 64×64} \\
\cellcolor[rgb]{.95,.95,.95}Defenses$\downarrow$Models$\rightarrow$ & \cellcolor[rgb]{.95,.95,.95}RN18 & \cellcolor[rgb]{.95,.95,.95}RN50 & \cellcolor[rgb]{.95,.95,.95}DN121 & \cellcolor[rgb]{.95,.95,.95}MNV2 & \cellcolor[rgb]{.95,.95,.95}\textbf{AVG} & \cellcolor[rgb]{.95,.95,.95}RN18      & \cellcolor[rgb]{.95,.95,.95}RN50 &\cellcolor[rgb]{.95,.95,.95} DN121 & \cellcolor[rgb]{.95,.95,.95}MNV2 & \cellcolor[rgb]{.95,.95,.95}\textbf{AVG} \\ \midrule[1.8pt]
w/o & 25.74    & 26.66    & 21.70       & 16.30       & 22.60        & 39.7     & 49.8          & 28.8        & 31.5        & 37.5 \\
CO & 25.46    & 29.20    & 23.90       & 17.58       & 24.04        & 42.1     & \cellcolor[rgb]{0.686, 0.67, 0.67}\textbf{56.2} & 38.2        & 36.9        & 43.4         \\
MU & 34.96    & 19.38    & 27.78       & 15.60       & 24.43        & 23.1     & 50.5          & 43.0        & 30.5        & 36.8         \\
CM & 16.54    & 24.04    & 23.58       & 8.00        & 18.04        & 32.3     & 50.5          & 29.8        & 25.9        & 34.6         \\
ISS-G & 14.92    & 13.50    & 9.78        & 5.78        & 11.00        & 12.5     & 19.1          & 11.3        & 11.1        & 13.5         \\
AT & \cellcolor[rgb]{0.686, 0.67, 0.67}\textbf{37.82}    & 36.80    & 30.34       & 41.42       & 36.60        & 53.7     & 53.3          & 51.9        & 51.0        & 52.5         \\
ISS-J & 30.10    &\cellcolor[rgb]{0.686, 0.67, 0.67} \textbf{37.04}    & 25.52       & 28.04       & 30.18        & 43.9     & 49.1          & 44.2        & 43.0        & 45.1         \\
\textbf{COIN (Ours)}          	 	 	 	 
&37.80 	&35.38 	&\cellcolor[rgb]{0.686, 0.67, 0.67}\textbf{35.22} 	&\cellcolor[rgb]{0.686, 0.67, 0.67}\textbf{41.50}	&\cellcolor[rgb]{0.686, 0.67, 0.67}\textbf{37.48} & \cellcolor[rgb]{0.686, 0.67, 0.67}\textbf{62.7}     & 54.7          & \cellcolor[rgb]{0.686, 0.67, 0.67}\textbf{65.1}        & \cellcolor[rgb]{0.686, 0.67, 0.67}\textbf{61.0}        & \cellcolor[rgb]{0.686, 0.67, 0.67}\textbf{60.9}\\ \bottomrule[1.8pt]
\end{tabular}
}
\caption{\textbf{Defenses against CUDA with large datasets.} The \textit{test accuracy} (\%) results on ImageNet100 (224×224) and ImageNet20 (64×64).} 
\label{tab:imagenet} 
\vspace{-4.5mm}
\end{table}

\subsection{Comparison with SOTA Defenses}

We compare COIN with 10 SOTA defense schemes, \ie,  AT~\cite{bettersafe}, ISS~\cite{liu2023image} including JPEG compression (ISS-J), and grayscale transformation (ISS-G), AVATAR~\cite{dolatabadi2023devil}, OP~\cite{pedro2023whatcanwe}, AA~\cite{qin2023learning}, and ECLIPSE~\cite{wang2024eclipse}. We also apply four defenses, differential privacy SGD (DP-SGD)~\cite{hong2020effectiveness,zhang2021feddpgan}, cutmix (CM)~\cite{cutmix}, cutout (CO)~\cite{cutout}, and mixup (MU)~\cite{mixup} that are popularly used to test against UEs. 

We do not compare EPD with the UE detection method proposed by~\cite{zhu2024detection} because our settings differ. Zhu~\etal's detection scheme is used to distinguish between clean samples and unlearnable samples (unable to distinguish convolution-based UEs), whereas EPD is designed to differentiate convolutional samples from other samples.

\subsection{Evaluation Metrics}
For evaluation of the COIN's effectiveness, consistent with previous works~\cite{wang2024eclipse,wang2024pointapa}, 
we employ $test$ $accuracy$, \ie, the model accuracy on a clean test set after training on the datasets. 
The higher the $test$ $accuracy$, the better the defense effectiveness. 
For evaluation of the detection performance, similar to~\citet{liu2023detecting}, we employ \textit{binary classification accuracy} (ACC) (\%) and \textit{Area Under the Receiver Operating Characteristic  Curve} (AUC) to measure the performance, where ACC is calculated as: 
\begin{equation}
    ACC = \frac{TP+TN}{TP+TN+FP+FN}
\end{equation}
AUC is the area under the ROC curve, indicating the model's ability to distinguish between classes. Ranging from 0 to 1, a higher AUC signifies better detection performance.

\subsection{Evaluation of Defense Scheme COIN} 
The test accuracy (\%) results are presented in~\cref{tab:main,tab:imagenet,tab:vudahuda}. ``\textbf{AVG}" denotes the average value for each row.
The values covered by \colorbox[RGB]{175,171,171}{gray} denote the best defense effect. 
It can be observed the average performance of existing defenses against CUDA lag behind COIN, ranging from 19.17\% to 44.63\% as shown in~\cref{tab:main}. 
Meanwhile, COIN maintains an advantage of 0.88\%-26.48\% on the large dataset,  8.4\%-47.4\% on the medium dataset as shown in~\cref{tab:imagenet}.
The reason AT lags behind COIN can be deduced from  that the multiplicative noise underlying convolutional UEs is more easily defeated by the multiplicative random noise used by COIN, rather than the additive noise on which AT is based~\cite{pgd}. We further evaluate defenses against VUDA and HUDA, as shown in~\cref{tab:vudahuda}. 
It can also be seen that COIN is effective in defending against these convolution-based UEs and performs better than other defense solutions by 23.84\% to 40\%. 
Regarding practicality, while we acknowledge that COIN cannot be effective for all types of UE, it still shows effectiveness against some bounded UEs including URP~\cite{bettersafe}, OPS~\cite{wu2022one}, which improves test accuracy from 16.8\%, 28.4\% to 81.1\%, 80.1\%.

\begin{table}[t]
\centering
\scalebox{0.49}{\begin{tabular}{c|ccccc|ccccc}
\toprule[1.8pt]
\cellcolor[rgb]{.95,.95,.95} Datasets & \multicolumn{5}{c|}{\cellcolor[rgb]{.95,.95,.95} VUDA CIFAR-10}  & \multicolumn{5}{c}{\cellcolor[rgb]{.95,.95,.95} HUDA CIFAR-10} \\
\multicolumn{1}{l|}{\cellcolor[rgb]{.95,.95,.95}Defenses$\downarrow$Models$\rightarrow$} & \cellcolor[rgb]{.95,.95,.95}RN18       & \cellcolor[rgb]{.95,.95,.95}VGG16          & \cellcolor[rgb]{.95,.95,.95}DN121    & \cellcolor[rgb]{.95,.95,.95}MNV2    & \cellcolor[rgb]{.95,.95,.95}\textbf{AVG}            & \cellcolor[rgb]{.95,.95,.95}RN18       & \cellcolor[rgb]{.95,.95,.95}VGG16          & \cellcolor[rgb]{.95,.95,.95}DN121    & \cellcolor[rgb]{.95,.95,.95}MNV2    & \cellcolor[rgb]{.95,.95,.95}\textbf{AVG}            \\ \midrule[1.8pt]
w/o   & 40.36          & 43.95          & 44.03          & 36.86          & 41.30          & 39.65          & 36.63          & 50.32          & 34.73          & 40.53          \\
MU & 41.13          & 41.66          & 39.98          & 39.55          & 40.58          & 43.45          & 44.54          & 48.96          & 42.01          & 43.91          \\
CM & 42.15          & 37.71          & 39.62          & 37.95          & 39.36          & 38.56          & 39.68          & 50.38          & 41.37          & 41.87          \\
CO & 39.93          & 38.95          & 36.64          & 38.37          & 38.47          & 40.53          & 40.18          & 47.74          & 36.58          & 40.70          \\
DP-SGD & 39.39          & 41.01          & 46.04          & 37.26          & 40.93          & 36.76          & 41.08          & 46.34          & 35.75          & 40.17          \\
AVATAR & 39.25          & 38.68          & 41.64          & 39.37          & 39.74          & 43.53          & 39.84          & 52.10          & 46.28          & 42.27          \\
AA & 51.87          & 10.00          & 60.09          & 51.42          & 43.35          & 56.83          & 10.00          & 60.09          & 61.08          & 46.27          \\
OP & 42.67          & 39.26          & 46.25          & 41.04          & 42.31          & 45.36          & 46.77          & 59.80          & 51.79          & 49.21          \\
ISS-G & 38.50          & 32.88          & 38.46          & 33.66          & 35.88          & 22.43          & 43.08          & 29.33          & 34.52          & 33.05          \\
ISS-J & 51.49          & 45.00          & 45.54          & 38.57          & 45.15          & 49.89          & 49.48          & 42.23          & 40.58          & 45.47          \\
AT & 43.40          & 38.68          & 46.88          & 36.33          & 41.32          & 48.62          & 41.19          & 50.48          & 43.74          & 45.07          \\
ECLIPSE & 31.09          & 33.47          & 35.56          & 29.86          & 32.49          &  34.57         & 39.04          & 41.12          & 35.35          & 37.52         \\
\textbf{COIN (Ours)} &\cellcolor[rgb]{0.686, 0.67, 0.67}\textbf{74.98} & \cellcolor[rgb]{0.686, 0.67, 0.67}\textbf{68.74} & \cellcolor[rgb]{0.686, 0.67, 0.67}\textbf{74.96} & \cellcolor[rgb]{0.686, 0.67, 0.67}\textbf{71.07} & \cellcolor[rgb]{0.686, 0.67, 0.67}\textbf{72.44} & \cellcolor[rgb]{0.686, 0.67, 0.67}\textbf{75.18} & \cellcolor[rgb]{0.686, 0.67, 0.67}\textbf{71.75} & \cellcolor[rgb]{0.686, 0.67, 0.67}\textbf{75.78} & \cellcolor[rgb]{0.686, 0.67, 0.67}\textbf{70.10} & \cellcolor[rgb]{0.686, 0.67, 0.67}\textbf{73.05} \\ \bottomrule[1.8pt]
\end{tabular}}
\caption{\textbf{Defenses against VUDA and HUDA.} The \textit{test accuracy} (\%) results on CIFAR-10 dataset.} 
\label{tab:vudahuda} 
\vspace{-3mm}
\end{table}



\subsection{Evaluation of Detection Scheme EPD}
We formulate the detection scenario with a dataset (size of 50000) combining an even distribution of six UEs among 12 SOTA UEs CUDA~\cite{sadasivan2023cuda}, HUDA, VUDA, OPS~\cite{wu2022one}, AR~\cite{sandoval2022autoregressive}, URP~\cite{bettersafe}, UHP~\cite{bettersafe}, LSP~\cite{syn}, TAP~\cite{adv}, EM~\cite{unl}, EFP~\cite{wen2023adversarial}, and SEP~\cite{chen2022self}.
We set the following combinations for CIFAR-10:
\begin{itemize}
    \item $\mathcal{S}_1 \longrightarrow$ CUDA, VUDA, HUDA, OPS, AR, URP
    \item $\mathcal{S}_2 \longrightarrow$ CUDA, VUDA, HUDA, OPS, LSP, URP
    \item $\mathcal{S}_3 \longrightarrow$ CUDA, VUDA, HUDA, TAP, EM, EFP
    \item $\mathcal{S}_4 \longrightarrow$ CUDA, VUDA, HUDA, OPS, EFP, SEP,  
\end{itemize}
and the following is for ImageNet-20:
\begin{itemize}
    \item $\mathcal{S}_5 \longrightarrow$ CUDA, VUDA, HUDA, LSP, AR, URP
    \item $\mathcal{S}_6 \longrightarrow$ CUDA, VUDA, HUDA, EM, TAP, UHP
    \item $\mathcal{S}_7 \longrightarrow$ CUDA, VUDA, HUDA, TAP, UHP, AR.
\end{itemize}
The results of EPD on CIFAR-10 and ImageNet20 are shown in~\cref{tab:sample_detection}. 
It can be seen that EPD's ACC and AUC are both high, demonstrating excellent detection performance.

\begin{table}[t]
\scalebox{0.65}
{
\begin{tabular}{c|ccccc|cccc}
\toprule[1.8pt]
\cellcolor[rgb]{.95,.95,.95}Datasets & 
\multicolumn{5}{c|}{\cellcolor[rgb]{.95,.95,.95}CIFAR-10 UEs}  & 
\multicolumn{4}{c}{\cellcolor[rgb]{.95,.95,.95}ImageNet20 UEs} \\
\cellcolor[rgb]{.95,.95,.95}Settings & 
\cellcolor[rgb]{.95,.95,.95}$\mathcal{S}_1$ & 
\cellcolor[rgb]{.95,.95,.95}$\mathcal{S}_2$ & 
\cellcolor[rgb]{.95,.95,.95}$\mathcal{S}_3$ & 
\cellcolor[rgb]{.95,.95,.95}$\mathcal{S}_4$ & 
{\cellcolor[rgb]{.95,.95,.95}\textbf{AVG}} &
\cellcolor[rgb]{.95,.95,.95}$\mathcal{S}_5$ & 
\cellcolor[rgb]{.95,.95,.95}$\mathcal{S}_6$ & 
\cellcolor[rgb]{.95,.95,.95}$\mathcal{S}_7$ &   
{\cellcolor[rgb]{.95,.95,.95}\textbf{AVG}} \\ \midrule[1.8pt]
ACC      &90.40 	&90.94 	&87.89 	&87.63 	&\textbf{89.21} 
& 99.92	&99.93	&99.89	&\textbf{99.91}                \\
AUC      &0.904 	&0.909 	&0.878 	&0.876 	&\textbf{0.891}               &0.999	&0.999	&0.998	&\textbf{0.998}               \\ \bottomrule[1.8pt]
\end{tabular}}
\caption{\textbf{Evaluation of EPD.} The ACC (\%) and AUC ‌results  with diverse settings on CIFAR-10 and ImageNet datasets.} 
\label{tab:sample_detection}
\vspace{-5mm}
\end{table}

\subsection{Hyperparameter Analysis }
\begin{figure}[t]
\centering
\scalebox{0.22}{  
\includegraphics[width=\textwidth]{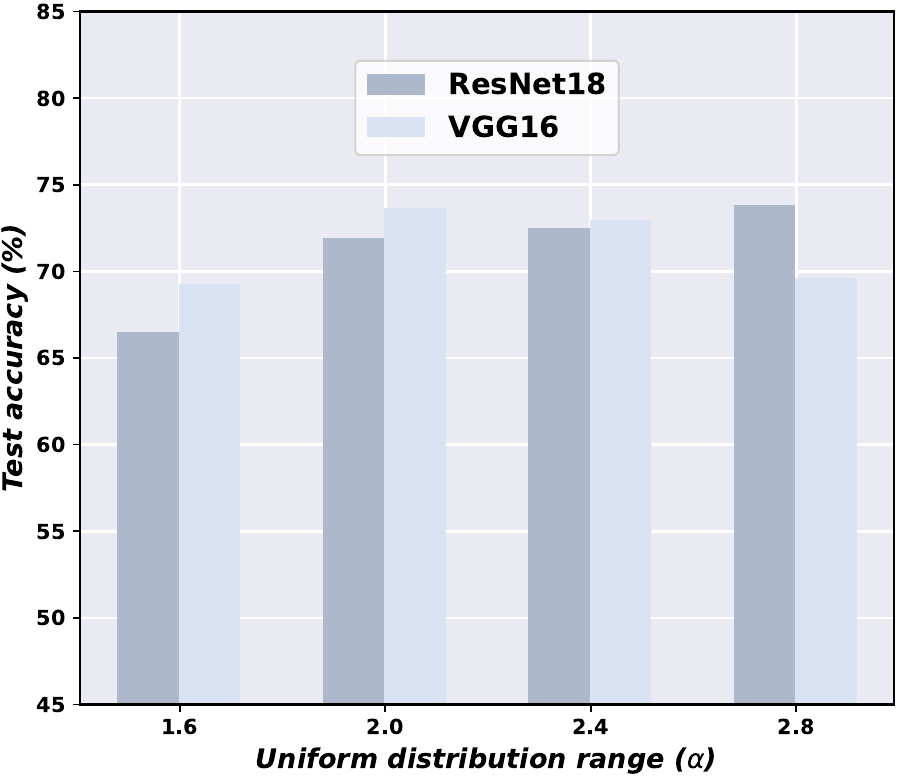}}
\scalebox{0.24}{  
\includegraphics[width=\textwidth]{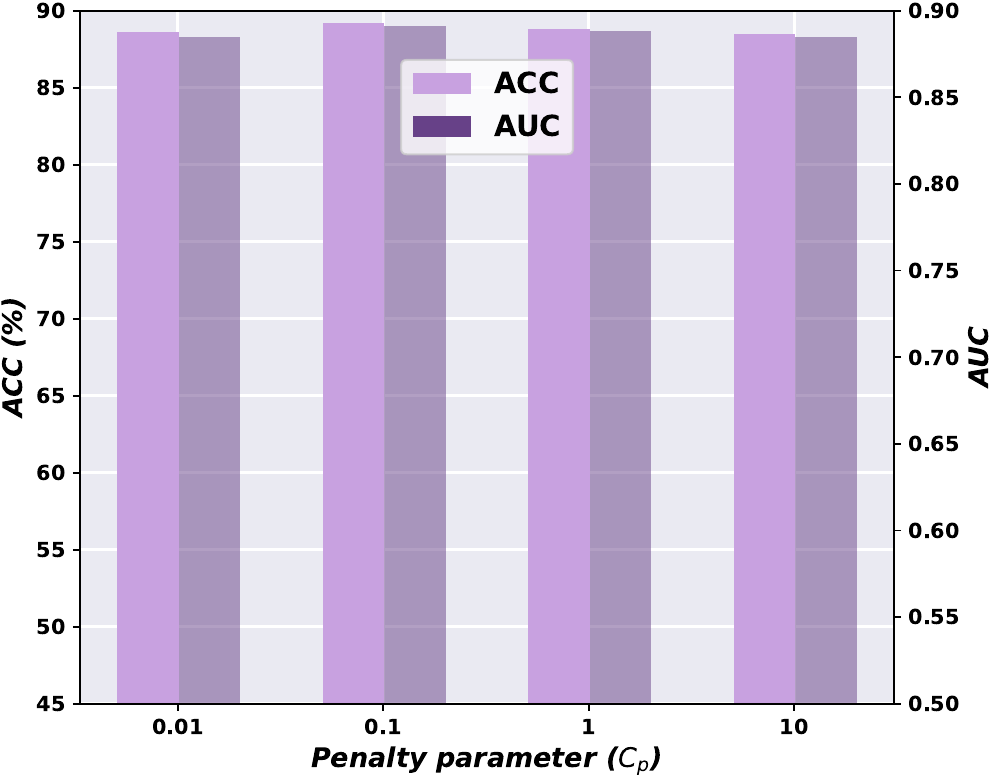}}
\caption{\textbf{Hyper-parameter analysis:} (a) The impact of $\alpha$ on the effectiveness of COIN against CUDA. (b) The impact of penalty parameter $C_p$ on the average effectiveness of EPD across from $\mathcal{S}_1$ to $\mathcal{S}_4$. }
\label{fig:ablation_alpha_time}
\end{figure}

We investigate the impact of $\alpha$ on the effectiveness of COIN, as shown in~\cref{fig:ablation_alpha_time} (a). 
The test accuracy against CUDA increases initially with the rise in $\alpha$ and then starts to decrease. This is because initially, as $\alpha$ increases, the CUDA perturbations gradually become disrupted. 
However, as $\alpha$ continues to increase, excessive corruptions damage image features, compromising defense effect. 
We empirically opt for an $\alpha$ of 2.0 that yields the highest average defense effect.
Additionally, given that the penalty parameter $C_p$ is a key hyperparameter in SVM training for regulating overfitting and underfitting, we examine its effect on EPD performance as illustrated in~\cref{fig:ablation_alpha_time} (b).
It can be seen that the average performance is best when $C_p$ is set to 0.1, which is also the default parameter used in our experiments.

\section{Related Work}
\subsection{Unlearnable Examples}
Current UEs can be classified into  bounded UEs and convolution-based UEs. 
The bounded UE methods utilize shortcut-based approaches to generate additive unlearnable noise~\cite{unl,adv,fu2021robust,wu2022one}. 
Nonetheless, recent studies indicate that commonly used defenses like JPEG compression can address these bounded UEs~\cite{liu2023image}. 
Recently, ~\citeauthor{sadasivan2023cuda} propose a new type of convolution-based UE via class-wise convolutional filters to generate multiplicative noise without norm constraints (CUDA). 
Unfortunately, all currently available defense approaches prove ineffective against it, and no current research has explored viable defense strategies.
To expand the scope of convolution-based UEs, we propose two new types of convolution-based UEs via  vertical filters and horizontal filters, termed as VUDA and HUDA, revealing that these convolution-based UEs easily defeat existing defenses.

\subsection{Defenses Against UEs}
Tao~\etal~\shortcite{bettersafe} experimentally and theoretically demonstrate AT~\cite{pgd} can effectively defend against bounded UEs, and Liu~\etal~\shortcite{liu2023image} discover that even simple image transformation such as JPEG compression can effectively defend against UEs.  
Thereafter, Qin~\etal~\shortcite{qin2023learning} employ \textit{adversarial augmentations} (AA), Dolatabadi~\etal~\shortcite{dolatabadi2023devil} (AVATAR) and Wang~\etal~\shortcite{wang2024eclipse} (ECLIPSE) purify UEs through diffusion models~\cite{diffusion},  Segura~\etal~\shortcite{pedro2023whatcanwe} train a linear regression model to perform \textit{orthogonal projection} (OP), while Yu~\etal~\shortcite{yu2024purify} corrupt UEs via rate-constrained variational auto-encoders.  
Nevertheless, none of these defense approaches are tailor-made for convolution-based UEs, with all of them failing against them. 
As for detection, although Zhu~\etal~\shortcite{zhu2024detection} first propose a detection method for UEs, it is limited to bounded UEs and does not consider convolution types.



\section{Conclusion} 
In this work, we reveal that existing defenses are all ineffective against convolution-based UEs. 
To address this issue, we first propose an edge-pixel detection scheme to identify convolutional samples.
Besides, we model this process using multiplicative matrices applied to samples and propose two metrics, demonstrating that increasing these metrics mitigates convolution-based UEs, achieved through the design of a random matrix.
Based on this, we propose the first effective defense against convolution-based UEs, utilizing pixel-level random multiplication via interpolation. 
We further propose two convolution-based UEs, termed as VUDA and HUDA, to expand the scope of convolutional UEs. 
Extensive experiments verify the effectiveness of our proposed defenses against convolution-based UEs.  
 
{
\small 
\bibliography{aaai25}
}

\end{document}